# A comparative study on wearables and single-camera video for upper-limb out-of-the-lab activity recognition with different deep learning architectures.

Mario Martinez-Zarzuela[1], David González-Ortega[1], Míriam Antón-Rodríguez [1], Francisco Javier Díaz-Pernas[1], Henning Müller[2,3], Cristina Simon-Martinez[3]
[1] Department of Signal Theory and Communications and Telematics Engineering, University of Valladolid, Spain
[2]Medical faculty, University of Geneva, Switzerland.
[3]Institute of Informatics, University of Applied Sciences Western Switzerland (HES-SO) Valais-Wallis, Sierre, Switzerland

**Introduction:**
The use of a wide range of computer vision solutions, and more recently high-end Inertial Measurement Units (IMU) have become increasingly popular for assessing human physical activity in clinical and research settings [1]. Nevertheless, to increase the feasibility of patient tracking in out-of-the-lab settings, it is necessary to use a reduced number of devices for movement acquisition. Promising solutions in this context are IMU-based wearables and single camera systems [2]. Additionally, the development of machine learning systems able to recognize and digest clinically relevant data in-the-wild is needed, and therefore determining the ideal input to those is crucial [3].

**Research question:**
For upper-limb activity recognition out-of-the-lab, do wearables or single camera offer better performance?

**Methods**
Recordings from 16 healthy subjects performing 8 upper-limb activities from the VIDIMU dataset [4] were used. For **wearable** recordings, the subjects wore 5 IMU-based wearables and adopted a neutral pose (N-pose) for calibration. Joint angles were estimated with inverse kinematics algorithms in *OpenSense* [5]. Single-camera v**ideo** recordings occurred simultaneously. Single-camera v**ideo** recordings occurred simultaneously, and the subject's pose was estimated with *DeepStream* [6]. We compared various Deep Learning architectures (DNN, CNN, CNN-LSTM, LSTM-CNN, LSTM, LSTM-AE) for recognizing daily living activities. The input to the different neural architectures consisted in a 2-second time series containing the estimated joint angles and their 2D FFT. Every network was trained using 2 subjects for validation, a batch size of 20, Adam as the optimizer, and combining early stopping and other regularization techniques. Performance metrics were extracted from 4-fold cross-validation experiments.

**Results:**
In all neural networks, performance was higher with IMU-based wearables data compared to video. The best network was an LSTM AutoEncoder (6 layers, 700K parameters; wearable data accuracy:0.985, F1-score:0.936 (Fig. 1); video data accuracy:0.962, F1-score:0.842). Remarkably, when using video as input there were no significant differences in the performance metrics obtained among different architectures. On the contrary, the F1 scores using IMU data varied significantly (DNN: 0.849, CNN: 0.889, CNN-LSTM: 0.879, LSTM-CNN: 0.904, LSTM: 0.920, LSTM-AE: 0.936).

**Discussion:**

Wearables and video present advantages and disadvantages. While IMUs can provide accurate information about the orientation and acceleration of body parts, body-to-segment calibration and drift can affect data reliability. Similarly, a single camera can easily track the position of different body joints, but the recorded data does not yet reliably represent the movement with all degrees of freedom. Our experiments confirm that despite the current limitations of wearables, with a very simple N-pose calibration, IMU data provides more discriminative features for upper-limb activity recognition. Our results are consistent with previous studies that have shown the advantages of IMUs for movement recognition [7]. In the future, we will estimate how these data compare to gold-standard systems.

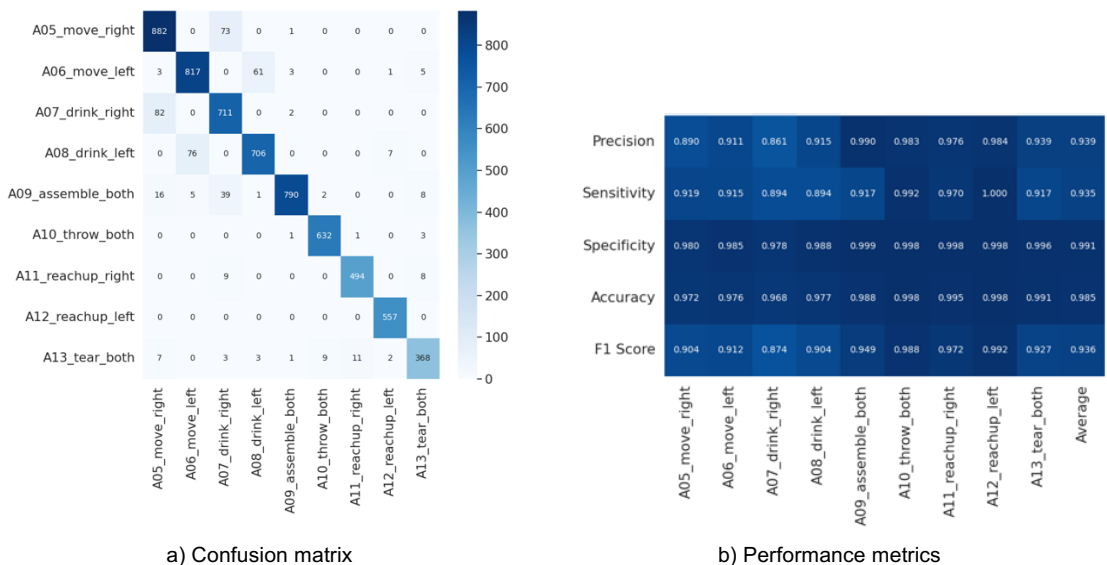

a) Confusion matrix        b) Performance metrics
**Fig 1**. Performance of upper-limb movement recognition using an LSTM-AE neural network and joint angles estimated through inverse kinematics from raw IMU data.